\documentclass[letterpaper, 10 pt, conference]{ieeeconf}  % Comment this line out if you need a4paper

\IEEEoverridecommandlockouts                              % This command is only needed if 
\usepackage{graphics} % for pdf, bitmapped graphics files
\usepackage{graphicx}
\usepackage{mathptmx} % assumes new font selection scheme installed
\usepackage{amsmath} % assumes amsmath package installed
\usepackage{amssymb}  % assumes amsmath package installed
\usepackage{cite}
\usepackage{pifont}
\usepackage[flushleft]{threeparttable}
\usepackage{array,booktabs,makecell}
\usepackage{multirow}
\DeclareMathAlphabet{\mathcal}{OMS}{cmsy}{m}{n}

\title{\LARGE \bf
PyroTrack: Belief-Based Deep Reinforcement Learning Path Planning for Aerial Wildfire Monitoring in Partially Observable Environments
}

\author{
    Sahand Khoshdel$^{1}$,
    Qi Luo$^{2}$,
    Fatemeh Afghah$^{1}$
\thanks{*This material is based upon work supported by the Air Force Office of Scientific Research under award number FA9550-20-1-0090 and the National Science Foundation under Grant Numbers CNS-2232048, and CNS-2204445.}
\thanks{$^{1}$Holcombe Department of Electrical and Computer Engineering, Clemson University, Clemson, SC, USA,
        {\tt\small \{skhoshd,fafghah\}@clemson.edu}}
\thanks{$^{2}$Department of Industrial Engineering, Clemson University, Clemson, SC, USA,
        {\tt\small qluo2@clemson.edu}}%
}

\usepackage{xcolor}

\renewcommand{\baselinestretch}{0.9}
\begin{document}

\maketitle
\thispagestyle{empty}
\pagestyle{empty}

%%%%%%%%%%%%%%%%%%%%%%%%%%%%%%%%%%%%%%%%%%%%%%%%%%%%%%%%%%%%%%%%%%%%%%%%%%%%%%%%
\begin{abstract}
%\red{Utilizing autonomous unmanned aerial vehicles (UAVs) in disaster management has gained more popularity over the past decade due to their numerous benefits, such as flexible mobility, real-time onboard processing, and low-risk operation compared to human-operated management systems.} 
Motivated by agility, 3D mobility, and low-risk operation compared to human-operated management systems of autonomous unmanned aerial vehicles (UAVs), this work studies  UAV-based active wildfire monitoring where a UAV detects fire incidents in remote areas and tracks the fire frontline. A UAV path planning solution is proposed considering realistic wildfire management missions, where a single low-altitude drone with limited power and flight time is available. Noting the limited field of view of commercial low-altitude UAVs, the problem formulates as a partially observable Markov decision process (POMDP), in which wildfire progression outside the field of view causes inaccurate state representation that prevents the UAV from finding the optimal path to track the fire front in limited time. Common deep reinforcement learning (DRL)-based trajectory planning solutions require diverse drone-recorded wildfire data to generalize pre-trained models to real-time systems, which is not currently available at a diverse and standard scale.  To narrow down the gap caused by partial observability  in the space of possible policies, a belief-based state representation with broad, extensive simulated data is proposed where the beliefs (i.e., ignition probabilities of different grid areas) are updated using a Bayesian framework for the cells within the field of view. The performance of the proposed solution in terms of the ratio of detected fire cells and monitored ignited area (MIA) is evaluated in a complex fire scenario with multiple rapidly growing fire batches, indicating that the belief state representation outperforms the observation state representation both in fire coverage and the distance to fire frontline.
% \red{Following this framework, a spatiotemporal inference model is proposed on the observation map while encoding the uncertainty of knowledge with regards to partial observability and dynamics, and directly outputting a belief map. While dealing with sparse updates through masked back-propagation through time, the proposed belief-based inference model demonstrates improved state representation and adaptive policies compared to observation maps, especially in narrower views and faster fire scenarios.}

\end{abstract}

%%%%%%%%%%%%%%%%%%%%%%%%%%%%%%%%%%%%%%%%%%%%%%%%%%%%%%%%%%%%%%%%%%%%%%%%%%%%%%%%
\section{INTRODUCTION}
% \red{ A study from \cite{burke2021changing} shows that wildfires are been accountable for more than \%25 (up to half in western regions) of $PM_{2.5}$ particles in the U.S., having a massive impact on the amount of smoke exposure, and severe degradation of public health. Moreover,} \fa{could be removed}

% In recent decades, the world has witnessed an alarming increase in the frequency and severity of wildfires, giving rise to significant threats to both human health and the ecosystem.  Economic costs that emerged from wildfires have increased dramatically, resulting in an estimated \$84.9 billion cost from 1980 to 2019 in the U.S. \cite{smith20202010}. As a result, there is an urgent need for wildfire management systems to accurately monitor the wildfire spread. UAVs equipped with advanced sensing technologies have offered many promising capabilities for high-resolution aerial imaging used in wildfire monitoring.  

The increase in wildfire frequency and severity in recent decades has significantly impacted human health and ecosystems. Economic costs of wildfire damage was approximately \$84.9 billion from 1980 to 2019 in the U.S. \cite{smith20202010}. This highlights the critical need for effective wildfire management systems. As a result, the early management of wildfires, including early detection, monitoring, modeling, and suppression has gained increasing attention. 

UAVs, equipped with advanced sensing
technologies have offered many promising capabilities for high-resolution aerial imaging, and have shown significant potential in wildfire detection tasks, such as creating labeled data sets for wildfire detection with smoke occlusion \cite{survey}. The authors in \cite{shamsoshoara2021aerial} and \cite{chen2022wildland} created a dual RGB/IR aerial dataset using UAVs. Compared to manned aerial systems, UAVs have transformed disaster management with shorter mission start-up time, improved robustness to smoke, chemicals, and heat, and ease of deployment. Despite the advantages of utilizing UAVs for wildfire detection, tracking the fire frontline actively through the early stages, known as \textit{wildfire monitoring}, remains a challenge.  In wildfire monitoring, models aim to solve an optimization problem with the objective of fire coverage and with respect to UAV's trajectory along the fire frontiers with limited batteries while preventing damage caused by heat and smoke. The optimized UAV trajectory consists of coordinates in a 2D space over a planning horizon. In wildfire management operations, the altitude of the UAVs is usually pre-determined for safety to avoid collision with other UAVs and aircraft. \\

% Solving the fire monitoring problem in a highly dynamic environment, such as a wildfire which follows complex spatial-temporal patterns influenced by vegetation, humidity, and wind, is computationally challenging. Several works formulated the trajectory optimization for fire monitoring in complex environments as partially observable Markov decision processes (POMDP). The partial observability is due to the large scale of the wildland fire is not fully observable from the low-altitude UAV's point of view. To solve a POMDP directly, the decision maker should be aware of the transition function of the environment states (dynamics), the observation function which maps the environment states to observations, and the reward function.
% However, not all information is available in wildfire monitoring but to be learned through the observations.\\

The complexity of wildfire environments, characterized by dynamic spatio-temporal patterns and influenced by factors like vegetation and wind, makes wildfire monitoring a computationally intensive task. Addressing this challenge, several studies have adopted a Partially Observable Markov Decision Process (POMDP) framework for trajectory optimization in wildfire monitoring, acknowledging the challenges posed by the vast scale of wildfires and the limited observability from low-altitude UAVs. Since model-based learning approaches require extensive pre-training data, developing an implicit representation of the environment's dynamics through a belief state could address this issue. Specifically, a belief-based approach can encapsulate information about the fire location and spreading behavior by performing Bayesian updates on belief states with previous observations, and thus substitute memory-mapping with perfect recalls.  This belief-based approach can also keep track of the interactions among unobserved factors that directly affect the state transitions and rewards that are explicitly difficult to learn. This easy adaptation to complex dynamics makes belief-based methods superior to methods such as MPC and Kalman filters which have been extensively studies in the early literature.\\

The current belief-based models for UAV path planning encode the observation uncertainty because of low resolution or limited field of view (FOV) by a surrogate history-dependent distribution. In other words, belief-based models tend to estimate the environment dynamics implicitly through learning a representation of the underlying dynamics matrix. This approach may be more computationally burdensome as the model has no masked focus on the observed area and does not take into account the age of collected information.  Previous POMDP-based fire monitoring models did not consider vegetation density, vegetation type, and realistic wind patterns influencing the fire spread, nor did they consider power limits, angle deviation, and the risk of overheating hardware. To the best of our knowledge, this is the first work to study UAV path planning in a realistic wildfire environment with various vegetation types and densities, considering the practical flight and power limitations of UAV and the age of observed information.\\

This paper proposes a belief-based DRL solution for UAV path planning to detect and monitor forest fires where low-altitude UAVs have a limited FoV of the environment. Our proposed method narrows the gap between observations and the full state by holding on to beliefs about cells outside the FOV. As a result, the agent will associate detection and monitoring rewards with the believed states of the environment and the UAV's current status. We demonstrate the effect of belief by comparing it to a purely memory-based observational representation in dynamic scenarios of the wildfire simulation. In summary, the contributions of the paper include the following:
\begin{itemize}
    \item  A multi-modal simulated wildfire framework (vegetation density and type, wind dynamics model, etc.)
    %\vspace{0.1cm}
    \item  Belief state solution to a PODMP for wildfire monitoring (frontline tracking) regarding physical constraints. 
    %\vspace{0.1cm}
    \item Uncertainty-aware state representation based on certainty map and age of information.
\end{itemize}

\section{Related Works}

Among the various methods for autonomous-UAV-based wildfire monitoring, this section highlights POMDP-based approaches. Some employ the FARSITE wildfire simulation model (\cite{finney2004landscape}, \cite{Finney_1998}), like \cite{pham2018distributed}, who used it in a distributed controller for monitoring dynamic wildfires with multiple UAVs, optimizing coverage relative to UAV altitude. The authors in \cite{lin2018kalman} and \cite{seraj2022multi} use FARSITE to develop a Kalman-based spread modeling framework as an estimate of where the fire front is propagating and optimize the trajectory of a UAV fleet based on the estimated state. \cite{giuseppi2021uav} uses cellular automata to model the wildfire spread and Voronoi tesselation to generate waypoints for a single UAV to follow along the fire frontline.

All the aforementioned works use control-based or optimization approaches for the wildfire monitoring problem. However, rapidly growing fires need more controlling flexibility and the fire dynamics are often unknown a priori. More recently, UAV-based fire monitoring are modeled as MDP/POMDP and solved by learning the dynamics of the environment explicitly or implicitly. Specifically, the state for which the UAV decides to take the optimal estimated action is represented as a belief over the true hidden state of the environment. RL methods are a powerful tool in such scenarios where the environment model is not available to agents \cite{Azar}.

\vspace{0.1cm}
In this vein, \cite{julian2019distributed} formulates the wildfire monitoring problem as a POMDP. THe wildfire model considers the fuel and ignition states of the cells. The fire-spreading effects are modeled as the ignition probabilities of non-burning cells dependent on the proximity to ignited cells. This environment model does not fully capture the variability of contributing factors in a spread such as wind, vegetation, etc. Their reward function consists of front-line proximity, ignition coverage, banking tight circles over ignited areas, and redundant observations of multiple aircraft. This reward design is comprehensive but lacks considering the limited fuel/battery of the aircraft and direct penalties for overheated units onboard the aircraft. Furthermore, for their evaluation scenario, only a few fire patterns including circular, T-shaped, and arced fire shapes are considered, overlooking general wildfire patterns and modeling forest fires. \\

 \cite{shobeiry2021uav} also models the wildfire tracking problem as a POMDP, while the environment and agent models include subsystems: the targets (fire fronts), the sensors (UAVs), and the tracker.   The sensor state encapsulates the location, speed, and heading angles of each UAV.  The target state models the 2D coordinates of the active flames. The tracker state, parameterized by a posterior mean vector and covariance matrix, aims to predict the locations of the fire fronts. 
 Their policies will determine the forward acceleration and bank angle of the UAV, given observations about the aggregated state of the fire fronts with measurement errors. The evaluation is done for three main scenarios of one, two, and three UAVs, plus an evaluation of the robustness to wind. Despite the innovative approach of \cite{shobeiry2021uav} and their extensive evaluation, wind disturbance is modeled with only adding a constant disturbance on the acceleration of the UAV and not on affecting the fire spreading.

\section{System Model}
This section describes the forest fire and agent models, respectively. 
%First, the forest fire model will be described in detail, followed by the UAV model for the state, action and observation, and finally the reward function design will be described. 
A summary of the environment and agent parameters and notation is shown in Table I. 
% \green{ Given the complexity of wildfires, the models that cannot be optimized in a short amount of time, and need many data samples are impractical to be deployed onboard UAVs with limited flight time and sensing capabilities.
% As a result, an increasing tendency towards pre-trained models with augmented data, specifically simulation-generated data is observed. } \fa{it does not belong here. can be moved to related works where you talk about simulation- based models}

\subsection{\textbf{Forest Wildfire Model}}
We consider a forest environment as a $N \times N$ grid being monitored by a single low-altitude drone. The trajectory of the UAV is defined as direct paths between cell centers. The frames captured by the UAV's camera along the path from one cell to the other are discarded for further processing to reduce the computational burden. %(Only frames at the sampling times are considered)

% ********************* Begin of Figure ********************* %
% \label{figure:grid_env}
% \begin{figure}[h]
%     \centering  \centerline{\includegraphics[width=0.5\textwidth]{Images/Grid.png}}
%     \caption{UAV model for fire-front tracking \fa{the figure is not easily readable. Let's show the wind vector and neighbor cells here too}}
% \end{figure}
% ********************* End of Figure *********************** %

\begin{table}
\label{tab:symbols}
  \caption{Parameter and Symbols Description} \label{table1}
  \centering 
  \begin{threeparttable}
    \begin{tabular}{cccc}
    %{m{15mm} m{70mm} m{18mm}}
    Envir. Params.  & Symbol & Agent Params. & Symbol\\
     \midrule\midrule
    Total State  & $\mathcal{S}_{env}$  &    Classification Error & $\mathcal{E}$   \\%new row
    \cmidrule(lr){1-2}\cmidrule(lr){3-4} Ignition Status & $F$ &  State  & $\mathcal{S}$  \\ 
    \cmidrule(lr){1-2}\cmidrule(lr){3-4} Remaining Fuel & $f$ &  Orientation & $\phi_{U}$\\ 
    \cmidrule(lr){1-2}\cmidrule(lr){3-4} Wind Magnitude & $A$  & Belief & $b$ \\ 
    \cmidrule(lr){1-2}\cmidrule(lr){3-4} Wind Direction & $\phi$ &  Certainty Factor & $c$  \\
    \cmidrule(lr){1-2}\cmidrule(lr){3-4} Parallel Wind Magnitude & $W_{||}$ &  Battery & $P_{U}$ \\
    \cmidrule(lr){1-2}\cmidrule(lr){3-4} Vegetation Density & $\rho$ &   Action  & $\mathcal{A}$ \\ 
    \cmidrule(lr){1-2}\cmidrule(lr){3-4} Vegetation Type  & $V$ & Reward & $\mathcal{R}$\\ 
    \cmidrule(lr){1-2}\cmidrule(lr){3-4} Vegetation Radius & $R_{v}$ &  Observation & $\mathcal{Z}$ \\   
    \cmidrule(lr){1-2}\cmidrule(lr){3-4} Cell Neighborhood & $N_{i,j}, \epsilon_{r}$ &  Policy & $\mathcal{\pi}$  \\   

    \midrule\midrule
    \end{tabular}
    
\end{threeparttable}
\end{table}

\subsubsection{\textbf{Environment State Parameters}}

 The state of each of the cells within the grid at time $t$ is represented by  $\mathcal{S}_{env}^{t}(i, j)=\{F^{t}_{ij},f^{t}_{ij},W^{t}_{ij}, \rho_{ij}, V_{ij}\}$, listed as follows:

\vspace{0.2cm}
\noindent\textbf{Ignition State} The ignition indicator, $F^{t}_{ij}=\left \{0, 1, 2\right \}$, represents the 'not-ignited', 'ignited', and 'burned-out' states, respectively. \\
    
\noindent \textbf{Remaining Fuel} The remaining fuel within the cell, $f^t_{ij}$  controls the fire intensity within the cell. Ranging from an initial value $f^0_{ij} = kf_{0}$, descending down to 0, after the cell finishes all the fuel within the cell is burnt out. The scale factor $k$ is defined as proportional to the vegetation density. The spread of the fire from a point within the cell to the corners is the factor controlling the probability of spread to the adjacent cells. However, for the sake of simplicity, we treat the remaining fuel as a co-variate of the progression of a spot fire within the cell. \\

\noindent \textbf{Vegetation Density Level} The density level, $\rho_{ij}$, corresponds to the amount of initial fuel available in a cell. In this work, we consider 5 different vegetation levels,  ($\rho_{ij} = k\rho_{0};\;k=\left\{1, ..., 5\right\})$ modeling various density levels present in the environment.\\

\noindent \textbf{Vegetation Type} The vegetation type, $V_{ij}$ controls how fast a cell finishes its fuel. \\

\noindent \textbf{Wind Magnitude and Phase} The magnitude and phase of the local wind around cell $(i,j)$, $W^{t}_{ij}=\left \{ A^{t}_{ij}, \phi^{t}_{ij} \right \}$, directly affects the spread probability. The temporal and spatial pattern of the wind's magnitude and phase are described in the next section $(\,\vec{W}(x, y, t) = A(x, y, t)\,e^{j\phi(x, y, t)}\,)$.  

\vspace{0.3cm}
\subsubsection{\textbf{Environment State Initialization}}
To generate the simulated data, we first generate $N_v$ random circular vegetation patches inside the grid environment, with a radius $R_v$ in which: ($R_{v}^{min} \leq R_v\leq R_{v}^{max}$). The $k^{\text{th}}$ patch has an assigned discrete vegetation density $\rho^{k}_{v} = \{1, ..., 5\}$ and vegetation type $V_{k} = \{1, ..., 5\}$, modeling the consumption rate of the fuel material. The vegetation density and type of every cell within this patch ($V_{ij}$) is set to $V_{k}$ and considered constant across the progression simulation. The initial spot fires are chosen randomly within the vegetated regions. Finally, the wind magnitude and phase are initialized across the grid based on arbitrary patterns, some of which are formulated in Eq. \ref{equation: initial_wind_phase}. To avoid the complexity of the empirical fitted distributions and CFD simulator solutions, a simple spatial and temporal decomposition is considered such that for the wind phase, different cells follow the same temporal pattern but are different in a lag/lead phase, relative to each other. $(\phi(x, y, t) = \Delta\phi(t) + \phi_{\,0}(x, y))$. \\
% Fig. \ref{figure:wind_init_phase} represents some sample phase and magnitude initialization for the wind phase
\vspace{-0.2cm}
\begin{equation}
\label{equation: initial_wind_phase}
\begin{aligned}
\phi_{\,0}(x, y) &= \{\frac{x\pi}{m_{\phi}N},\, 
\frac{y\pi}{m_{\phi}N},\, tg^{-1}\,(\frac{y-\frac{N}{2}}{x-\frac{N}{2}}),\, ...\}\;\;;m_\phi\geq\frac{1}{2} 
\end{aligned}
\end{equation}
where $m_{\phi}$ represents the spatial spread factor in the phase component. The lower bound for $m_\phi$ is to ensure a spatially-unique phase pattern induced by an initial phase limitation $({0}\leq\;\phi_{0}(x, y)\;\leq2\pi)$. \\

% \begin{figure}
%     \centering
%      \centerline{\includegraphics[width=0.4\textwidth]{sample_init_phase.jpg}}
%      \caption{Sample Initialization of Wind Phase - From left to right: vertically/horizontally varying, radially inward/outward}
% \label{figure:wind_init_phase}
% \end{figure}

The wind magnitude follows the same spatial and temporal decomposition ($(A(x, y, t) = \Delta{A}(t) + A_{\,0}(x, y))$). For the initial wind magnitude, due to the higher turbulence around fire centers, we consider a 2D Gaussian radial basis function (GRBF) around every ignited cell and choose the magnitude of every cell based on the RBF for which the cell is closer to its center. (Eq. \ref{equation: initial_wind_mag})
\begin{equation}
\label{equation: initial_wind_mag}
\begin{aligned}
A_{0}(x, y) &= A_{max}\;exp[\,(\frac{-1}{2\sigma_{rad}^2}) \underset{ij\in{\,I}}{min}\;d_{xy,ij}^2\,]\\ \overset{(\epsilon_{rad}= 3\sigma_{rad})}\Rightarrow  &= A_{max}\;exp[-(\frac{9}{2\epsilon_{rad}^2})\underset{ij\in{\,I}}{min}\;d_{xy,ij}^2]
\end{aligned}
\end{equation}

\begin{figure}
    \centering
     \centerline{\includegraphics[width=0.5\textwidth]{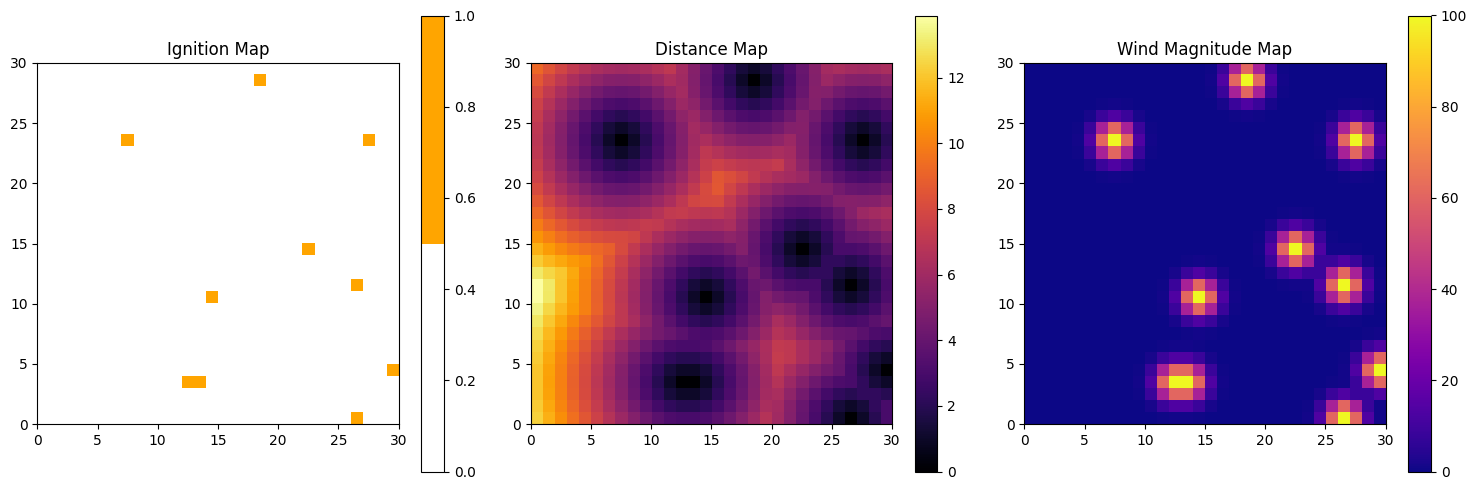}}
     \caption{\textbf{Sample Initialization of Wind Magnitude for $N = 30,\;N_{ign} = 10,\;A_{max} = 100,\;\epsilon_{rad} = 3$. $N_{ign}$} represents the number of initial ignitions. Maps: (Left: Ignition State, Middle: Distance from Fire, Right: Wind Magnitude)}
     % \fa{1) use bold font and define the subfigures in the caption. 2) I don't see the definition of $N_{ign}$ in the text}}
\label{figure:wind_init_mag}
\end{figure}

In Eq. \ref{equation: initial_wind_mag}, $I$ is the set of initially ignited cells, $N$ represents the grid size, $d_{xy,i'j'}$ is the distance to the source cell,  $\epsilon_{rad}$ is a cell radius for which the source cells RBF value is approximately zero. An initial configuration for the wind magnitude with noted parameters is displayed in Fig. \ref{figure:wind_init_mag}. 

% \fa{where was $\epsilon_{rad}$ was defined and why it's equal to three?} \fa{the size of grid was previously defined n by n (first line of section Q, but here you use N to show the size. }\\
% \fa{why using i' and j' here and not i and j?}

\vspace{0.3cm}

\subsubsection{\textbf{Environment State Dynamics}}

The vegetation type and density are considered constant for every cell throughout the wildfire, ignoring vegetation departure and characteristic variation in a short monitoring period. The other variables including the remaining fuel, the wind's magnitude and direction vary over time, resulting in the ignition state transitions.  
\vspace{0.1cm}

\noindent{\textbf{Fuel Consumption.}}
A transition from ignition state '1' to '2' (burnout) happens when the fuel inside a cell finishes.  Eq. \ref{equation: Fuel_Consumption} shows the vegetation density scales the initial amount of fuel for a cell, while the vegetation type controls the fuel consumption rate. It should be noted that the fuel consumption rate, in reality, depends on many factors other than the fuel, of which the most important are heat and oxygen density in the surrounding area. For the sake of simplicity, we consider the wind intensity as a rough measure of the oxygen density in the cell. Finally, as the cell fuel runs out, the ignition state of the cell changes to '2', indicating a burnt cell. ( see (\ref{equation: Burnout}))
\begin{align}
    f_{ij}^{\;t}= \rho_{ij}\;exp(-\,V_{ij}(\,\frac{A_{ij}^{t}}{\max\,(A_{ij}^{t})}\;t) \label{equation: Fuel_Consumption} \\
    (F_{ij}^{t+1}| F_{ij}^{t} = 1)  = \begin{Bmatrix}
    2 & ;\;(f_{ij}^{t+1}= 0)\\ 
    1 & ;\;\;\;\;\;\;\;\;O.W.
    \end{Bmatrix} \label{equation: Burnout}
\end{align}

% \fa{do you mean to present the remianing fuel  $f_{ij}^t$ or the fuel consumption function$f_{ij}(t)$? }

\noindent{\textbf{Wind Dynamics.}}
Accurate simulation needs CFD modeling with multi-physics software, which is beyond the scope of this article. Therefore, we consider a simple sinusoidal temporal pattern to govern the phase and magnitude dynamics according to Eq. \ref{equation: wind_update}
\begin{equation}
\label{equation: wind_update}
\Delta\phi(t) = (\frac{\pi\,t}{T_{p}})
;\;\Delta{A}\,(t) = A_{b}sin(\frac{\pi\,t}{T_{m}}) 
\end{equation}

Here, $A_{b},\, T_{m},\, T_{p}$ stand for the base wind magnitude level, magnitude variation period, and phase variation period. 
% Fig. \ref{fig:wind_dynamics_1d} shows how the temporal components of magnitude and phase model the wind dynamics 

% \begin{figure}
%     \centering
%     \includegraphics[width=0.75\linewidth]{wind_dynamics_1d.jpg}
%     \caption{The wind magnitude and phase variation for $A_{max} = 100,\,A_{t} = 20,\,T_{m} = 20,\,\phi_{0} = \frac{\pi}{3},\, T_{p} =  100 $}
%     \label{fig:wind_dynamics_1d}
% \end{figure}

\noindent{\textbf{Ignition State.}}
% The ignition state changes from '0' to '1' when the fire spreads to a cell from its neighbor cells. It is assumed that the cells can only get ignited through the fire spread and no new cell is set on fire throughout the monitoring process (e.g., via lightning). The probability of a fire getting spread from an ignited cell $(i,j)$ as the source ($S_{ij,i'j'}^{t} = 1$), to a neighboring cell $(i',j')$  is shown in Eq. \ref{equation: SpreadingFire}  as a function of various factors, including the distance between the two cells ($d_{ij, i'j'}$), the remaining fuel of the source cell ($f_{ij}^t$), and the projection of the wind intensity parallel to the interconnecting line of the two cells, $W_{||ij}^{t}$, which is calculated based on the magnitude $A_{ij}^{t}$ and phase $\phi^{t}_{ij}$ of the source cell wind, and the angle of the interconnecting line). The aforementioned dependencies are formulated as three factors multiplied by one another ($F_{adj}, F_{fuel}, F_{wind}$), as described in Eq. (\ref{equation: SpreadingFire2}). To model no-wind scenarios, the wind factor is modified and normalized by a factor of 2. This way the other two factors are still responsible for causing ignitions.  Moreover, the adjacency factor formula, $\sigma_{spr}$, indicates the radius of impact as a likelihood factor for a cell to capture the fire from an ignited cell where a simple spatial Gaussian model is considered. In the rest of the paper,\,$d_{ij, i'j'}$ represents the distance between a sample cell $(i, j)$ with another cell $(i', j')$ which can be euclidean or non-euclidean.)

Fire spread is modeled by the transition of  the ignition state from '0' to '1' for cells next to currently ignited cells. The probability of spread from a source cell $(i,j)$ to an adjacent cell $(i',j')$, described as $S_{ij,i'j'}^{t} = 1$, depends on factors such as inter-cell distance $(d_{ij, i'j'})$, source cell fuel $(f_{ij}^t)$, and wind intensity $(W_{||ij}^{t})$, calculated using wind magnitude $(A_{ij}^{t})$, phase $(\phi^{t}{ij})$, and wind alignment angle, as shown in Eq. \ref{equation: SpreadingFire}. These factors are combined as $F{adj}, F_{fuel}, F_{wind}$ (Eq. \ref{equation: SpreadingFire2}), with the wind factor adjusted in no-wind scenarios. The impact radius, $\sigma_{spr}$, gauges the fire's spread likelihood, and throughout the paper, $d_{ij, i'j'}$ denotes the distance between cells

\begin{equation}
\label{equation: SpreadingFire}
\begin{aligned}
&p(F_{i'j'}^{t+1} = 1 | F_{i'j'}^{t} = 0) = \\ &\sum_{ij}p(F_{i'j'}^{t+1} = 1 | F_{i'j'}^{t} = 0,\,S_{ij,i'j'}^{t} = 1)p(S_{ij,i'j'}^{t} = 1)
\end{aligned}
\end{equation}
\begin{equation}
\label{equation: SpreadingFire2}
\begin{aligned}
p(F_{i'j'}^{t+1} &= 1 | F_{i'j'}^{t} = 0,\,S_{ij,i'j'}^{t} = 1) = \\
& \underbrace{(\frac{1}{2\sigma_{spr}^2}e^{-\frac{d_{ij,i'j'}^2}{2\sigma_{spr}^2}})}_{F_{adj}} \underbrace{\left(1 - \frac{f_{ij}^t}{f_{ij}^0}\right)}_{F_{fuel}} \left(\frac{1}{2}(1+ \underbrace{\frac{W_{||ij}^{t}}{\max\,A_{ij}^{t}})}_{F_{wind}}\right) \\ \\
W_{||ij}^{t} & = A_{ij}^{t}\,cos(|\theta_{ij,i'j'} - \phi^{t}_{ij}|)
\end{aligned}
\end{equation}
\begin{align}
\label{equation: SpreadingFire3}
& p(S_{ij,i'j'}^{t} = 1) = \frac{e^{-d_{ij, i'j'}^2}}{\sum_{ij}e^{-d_{ij, i'j'}^2}}
\end{align} 

\begin{figure}[h]
    \centering
    \centerline{\includegraphics[width=0.5\textwidth]{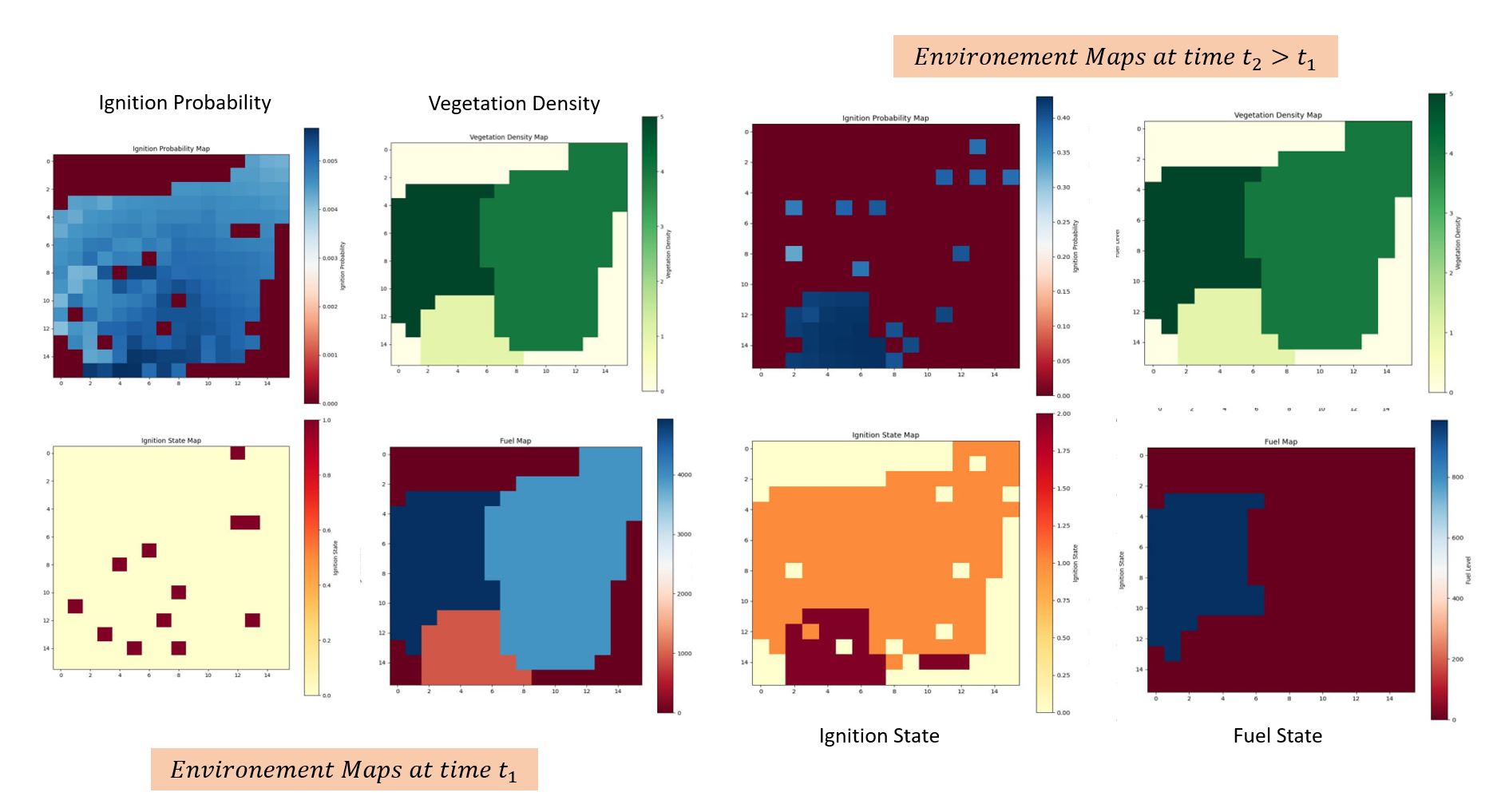}}
    \caption{The effect of fuel density and type on fuel consumption shown in a sample spread scenario. The denser vegetation patches have a higher initial fuel leading to a later burnout.}
    
    % \fa{not readable at all, enlarge the figs, use larger bold fonts and put a subfigure label and refer to them based on subfigure label in the capture}
    
    \label{fig:spread_scenario}
\end{figure}

The Fig. \ref{fig:spread_scenario} shows a simple scenario of the fire spread including ignition probability maps and fuel maps. Moreover, the effect of vegetation density on the initial amount of fuel within the cell and  the vegetation type on the burnout rate of a cell are seen, respectively.

% ********************* Begin of Figure ********************* %
% ********************* End of Figure *********************** %

\subsection{Agent Model}
\begin{figure}
\vspace{-10pt}
    \centering
    \includegraphics[width=\linewidth]{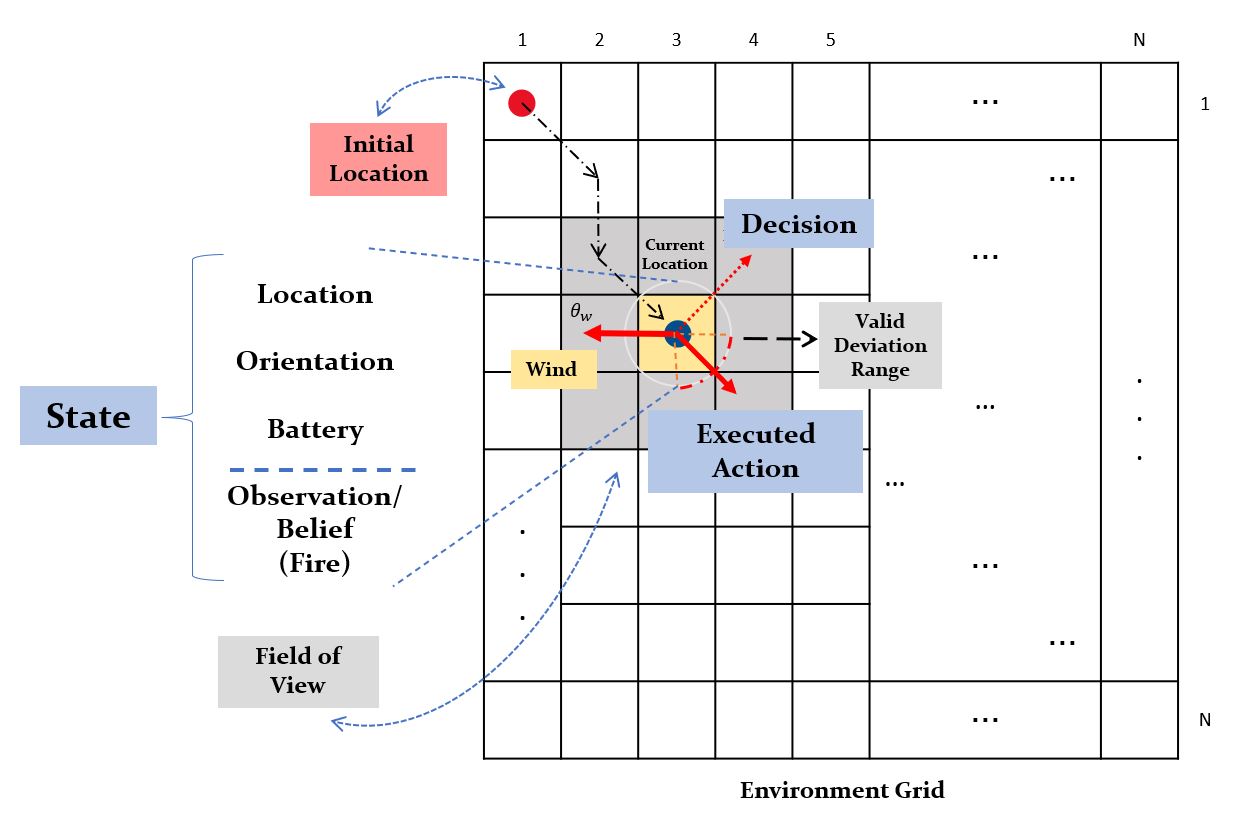}
    \caption{UAV model for fire-frontline tracking. The valid deviation denotes the acceptable action space which may differ from the action with highest value, thus the most optimal action in the valid range is selected.} 
    
    % \fa{colorcode the Fig to be more readable (use gray code for FoV text, red for initial location, use capital N to define the grid to follow the text notation. 2) what is the difference between decision and executed action (UAV is stable and no uncertainty related to how an action is executed (being impacted ) by Environment (wind). 3) use thicker lines for wind and direction. }
    
    \label{fig:Agent Model}
\end{figure}

 % To model the UAV as a reinforcement learning agent, a few assumptions should be considered. The altitude and speed of the UAV (in case of movement) is considered to be fixed throughout the mission, resulting in a fixed FOV (considerably small relative to the environment size). The agent and its interaction with the environment is modeled as Partially Observable MDP (POMDP) shown with the tuple  ($\mathbf{S},  \mathbf{A}, \mathbf{Z}, \mathbf{{T_{a}}^{ss'}}, \mathbf{{R_{a}}^{ss'}},\mathbf{{O^{a}_{s}}}, \mathbf{p_{s}^{0}}, \mathbf{\gamma})$ representing the state space, action space, observation space, transfer function, reward function, observation function, initial state probability distribution, and the discount factor. Solving a POMDP is more difficult due to the incomplete knowledge of the true state of the environment. The association of the state and state-action values to the rewards that depend on the full environment state introduces a new source of uncertainty. Hence, it is difficult to obtain a policy that maximizes the discounted reward over a sequence of observations. The sequence of states, observations, actions, and consequent rewards at every time step form a so-called history $\mathcal{H} = \{ s^1, a^1, a^1, s^2, a^2, a^2, ...\}$. Using this accumulated history is not memory efficient and thus the current state and observation of the agent is used (due to Markovity) to shape a policy $\pi^{a}_{s, z}$.

The UAV is modeled as a reinforcement learning agent within a POMDP framework, maintaining constant altitude and speed for a fixed FOV. The POMDP is represented by the tuple ($\mathbf{S}, \mathbf{A}, \mathbf{Z}, \mathbf{{T_{a}}^{ss'}}, \mathbf{{R_{a}}^{ss'}}, \mathbf{{O^{a}_{s}}}, \mathbf{p_{s}^{0}}, \mathbf{\gamma})$, covering state, action, and observation spaces, among others. Given the partial observability and complexity of environmental states, maximizing discounted rewards over observed sequences is challenging. The UAV's decision-making relies on current states and observations instead of inefficient history accumulation, following Markovity to form policy $\pi^{a}_{s, z}$.

 %  a Markov Decision Process (MDP). Each MDP consists of five main components. The state, the action, the transition probability, the reward function, and the discount factor ($\mathbf{S},  \mathbf{A}, \mathbf{{P_{a}}^{ss'}}, \mathbf{{R_{a}}^{ss'}}, \mathbf{\gamma}$). However, due to the limited field of view of the agent, the problem gets converted into a

\subsubsection{\textbf{State Space}}
% The state of the UAV consists of the coordinates of the current 2D location, the orientation direction, and the remaining battery level of the agent denoted by $S_{UAV} = \left \{(x, y), {\phi^{t}_{U}}, {P^{t}_{U}} \right \}$. The orientation limits the state-dependent action space and acts as a regulation to prevent choosing specific directions in some states, such as the edges of the grid, or as a limit on the deviation angle from the previous action. It should be noted that if the UAV executes the hovering action at a certain time step, there will be no limit on the deviation angle of the next time step.

The UAV's state is defined by its 2D coordinates, orientation, and battery level, expressed as $S_{UAV} = \left\{(x, y), {\phi^{t}_{U}}, {P^{t}_{U}}\right\}$. The orientation influences the available actions, specifically near grid edges or when adjusting the action's deviation angle. Notably, hovering actions remove the next step's deviation angle constraint.

\subsubsection{\textbf{Observation Space}}
% The observations of the agent in a POMDP, as the word itself suggests, are not complete, and thus do not reflect the true state of the environment at a given time.
% In other words, we can model the observation of the agent as a complex function of the environment's true state, and decompose every contributing factor to this erroneous observation. First of all, the UAV is only capable of observing one of the state parameters, the ignition state, and the hidden nature of the wind and the remaining fuel contribute towards partial observability even in a complete FOV. Beyond this point, the narrower the FOV is, the more outdated gathered information about an environment could potentially be.  Finally, the perceived ignition states are determined by the output of a classification module deployed onboard the UAV for real-time monitoring which causes computational limitations in terms of accuracy. Thus, many ignited cells can be misclassified as not-ignited/burnt and vice versa. Therefore, the observation of UAV at time $t$ is modeled as a classification error matrix multiplied by the true state of the cells within the FOV at time $t$. ($\mathbf{O^{t}_{i, j}} = \mathbf{\mathcal{E}(F^{t}_{i, j})}$

In a POMDP, the UAV's observations are incomplete, only capturing the ignition state and not reflecting the true environmental state, especially due to the hidden nature of wind and fuel levels. Additionally, the UAV's limited FOV leads to outdated information. Observed ignition states, affected by an onboard classification module, may inaccurately represent the actual states due to computational limitations. This is modeled as $\mathbf{O^{t}_{i, j}} = \mathbf{\mathcal{E}(F^{t}_{i, j})}$, where the classification error matrix ($\mathcal{E}$) impacts the observed state.

\subsubsection{\textbf{Action Space}}
% The action shows the direction that the UAV  planned to take in the next step ({denoted by $\theta^{t+1}$}). The action space is state-dependent as described above, but is a subset of the full action space. The full action space consists of 4 main and 4 diagonal directions for the drone to take in the next step, along with the action of choosing to hover above the cell, 
% $\mathcal{A}_H$ ($\mathcal{A} = \left \{ \frac{k\pi}{4}\;;\;\; k \in \left \{ 0, ..., 7 \right \}\right \} \cup \mathcal{A}_H $). Some works consider directions to be sampled from a continuous action space, but as the dynamicity of the environment and the partial observability already degrades the computational efficiency of achieving a sub-optimal trajectory, discrete action spaces are considered in this work.  Choosing a discrete action space  to reduce the complexity is a common practice among relevant works such as \cite{julian2019distributed} that consider a fixed bank angle deviation. (In Eq. \ref{equation: Action Space Limit}, $\mathcal{A}_s$ is the state-dependent action space which represents the limitation on actions that exit the grid or violate the deviation range.)
% \begin{equation}
% \label{equation: Action Space Limit}
% \theta^{t+1} \in \mathcal{A}_S \subseteq A \;\;\;\;\;if: \left \| \theta^{t+1} - \theta{t}\right \| < \Delta\theta_{dev} 
% \end{equation}

The UAV's action, $\theta^{t+1}$, indicates its next movement direction within a state-dependent action space. This includes 4 main and 4 diagonal directions, plus hovering ($\mathcal{A}_H$), collectively forming $\mathcal{A} = \left\{\frac{k\pi}{4}; k \in \left\{0, ..., 7\right\}\right\} \cup \mathcal{A}_H$. Given the dynamic, partially observable environment, we opt for discrete actions to enhance computational efficiency, aligning with practices in related works. The actions conform to grid constraints and deviation limits, as shown in Eq. \ref{equation: Action Space Limit}.
\begin{equation}
\label{equation: Action Space Limit}
\theta^{t+1} \in \mathcal{A}_S \subseteq A \;\;\;\;\;if: \left \| \theta^{t+1} - \theta{t}\right \| < \Delta\theta_{dev} 
\end{equation}

\subsection{Reward Function}
For the reward function, we have to model both the main task and the constraints in the reward function. Hence, the reward function is an aggregation of a set of sub-functions modeling the objectives and constraints. (Eq. \ref{equation: Agg. Reward})  $R_{obj}^{t}, R_{cstr}^{t}, R_{inf}^{t}$ represent the objective (main) reward, constrain penalty (power and battery), and information gain reward, respectively.
\begin{equation}
\label{equation: Agg. Reward}
 \mathbf{R^{t}_{total} = R_{obj}^{t} + R_{cstr}^{t} + R_{inf}^{t}} \;\ \\
\end{equation}

For this task, we consider detecting and tracking fires along the fire frontier as positive rewards representing the objective, while burnout (the UAV getting too close to the fire flame,) power consumption in movement/hovering, and the event of battery falling beyond a threshold (determining the recharge/return status) are modeled as negative rewards.
\begin{equation}
\begin{aligned}
\label{equation: Reward Obj.}
\mathbf{R_{obj}^{t}}  &= \alpha_{det}\frac{\left \| n_{det} \right \|}{\left \| n_{FoV} \right \|} + \alpha_{mon}e^{-d_{min}} ;\;\;\; \\ 
S_{det} &= \left \{ (i, j):\; F_{i, j}^{t} = 1, (i, j) \in S_{FoV} \right \}
\end{aligned}
\end{equation}
where $\alpha_{det}, \alpha_{mon}$ represent detection and monitoring reward coefficients, which aim to balance the two objectives of fire discovery and frontline tracking.

Conditions for which the negative rewards of battery depletion and burnout should be applied are shown with conditional identity functions. 
\begin{equation}
\label{equation: Reward Constrain}
\begin{aligned}
 \mathbf{R_{cstr}^{t}} &= \;\alpha_{mvm}R_{mvm}^{t} +
 \alpha_{P}R_{btr}\,\mathbf{1}(P^{\,t}_{U}< P^{\,\text{thr}}_{U}) \\&+\alpha_{brn}R_{brn}\,\mathbf{1}(F_{x,y}^{t} = 1)
\end{aligned}
\end{equation}
% \fa{please define all the notations, pretty please :)} 
To model the power consumption of physical movement accurately, we penalize the agent for moving $\gamma_m$ times more than hovering and also consider a wind-dependent coefficient $\beta_t$ which penalizes movement against the wind more than movement in its direction.

\begin{equation}
\label{equation:Reward Movement}
\begin{aligned}
 &R_{mvm}^{t} = \begin{Bmatrix}
 R_H&;\; A^t = A_H\\ 
 \gamma_{m}\;\beta^t R_M&;\; A^t  = \frac{k\pi}{4},  k \in \mathbb{N}
\end{Bmatrix} \\
& \beta^t = 1 - cos( A^t - \angle W^t)
\end{aligned}
\end{equation}
Moreover, in the case of using belief maps, we add a similarity reward that shows the accuracy of a belief about an area after observing it within the FOV.  
\begin{equation}
\label{equation: Reward Info.}
\begin{aligned}
 \mathbf{R_{inf}^{t}} &= - \alpha_{bel}I(b_{FoV}; z_{FoV}) \\
 I(X; Y) &= D_{KL} (P_{XY}(x, y) \left | \right | P_{X}(x)P_{Y}(y))
 \end{aligned}
\end{equation}
In Eq. \ref{equation: Reward Info.}, $I$ represents mutual information between two random variables (vectors), which itself is the Kullback-Leiber (KL) divergence of their joint probability density and the product of their marginal probabilities. After aggregating the aforementioned partial rewards ($R_{obj}^{t}, R_{cstr}^{t}, R_{inf}^{t}$), the aggregation weights $(\alpha_{det},\;\alpha_{mon},\; ...,\;\alpha_{bel},\;\gamma{m})$ should be hyper-tuned to obtain desirable results. Moreover, they can be adjusted dynamically through an episode to model the priority of the objectives or constraints in different phases of the mission. This dynamic focus helps the agent prevent the challenge of reward aggregation in multi-objective multi-constrained problems.

\section{Proposed Method}
In this section, the two mission phases of the UAV are described. Next, the DQN used for value estimation is discussed. Finally, the solution to the POMDP approach using belief maps, in terms of state representation, is explained. 

% As described before, due to the limited field of view of the agent, the problem gets formulated as a POMDP, with the variable full states, capturing the dynamics of the wildfire. These states are hidden to the agent, and the agent should either rely on its observations to infer these full states approximately. Here we propose a belief map, which keeps track of the environment map as the wildfire grows. The belief map is initialized by a Beta distribution which is the conjugate prior of a Bernoulli task. However, we only use this conjugate concept as an initialization point for the cells within the environment which evolve based on a Bernoulli trial. For updating the beliefs, we follow a similar path to the one that was taken for simulating the wildfire spread in the environment. Following this framework, the agent updates it's belief about the full state of the environment, based on the previous observations from a particular location which is not currently in the field of view of the agent. In particular, instead of centralizing GRBF functions on ignited cells, we slide a GRBF ‌kernel on the belief maps but only change the cells which have not previously been observed as ignited. The rate of belief updates are experimentally tuned to be synchronized with the average ignition rate of the environment. As applying a spatial kernel does not increase the computational burden significantly, in reality, this update can be done every few seconds as the drone flies around the frontier, to make sure the environment dynamics are captured with the highest resolution possible. 

\subsection{\textbf{\textit{Mission Phases}}}
Weight initialization is crucial in neural networks, including DQNs, due to the challenges posed by numerous local minimums in non-convex functions. Prior information plays a vital role, guiding the value network in the parameter space to be closer to true values. Our monitoring method adopts a 'Scan-and-Track' bi-phase approach, enabling the UAV to effectively initialize beliefs about the environment, countering the issue of initial ignorance of fire locations.

\subsubsection{\textbf{\textit{Scanning Phase}}} In the scanning phase, the agent calculates the shortest path along the whole environment based on its starting location, FOV size, and environment size, to create an initial fire grid map and update the wildfire model. This phase, conducted only in the first episode of each epoch, serves as DQN weight initialization. Here, the UAV follows a predetermined path, while enabling policy evaluation to enhance value estimates. 
 
\subsubsection{\textbf{\textit{Tracking Phase}}}
After one round of scanning the environment, we start the first episode of training by executing policies with an epsilon-greedy exploration-exploitation approach, where the epsilon is set to decay in a range of episodes in each epoch.

\subsection{\textbf{\textit{State Representation}}}
As discussed previously, the observations in a POMDP are a function of the true state and several sources of error in between. In this section, we will discuss different approaches toward state representation as inputs to value/policy networks in the wildfire monitoring problem. 

\subsubsection{\textbf{\textit{Observation-Based Representation}}}
As the UAV moves from one area to another, the observations become outdated and the observed state of a cell within the old area at a specific time $Z^t_{i, j}$ is no longer a good estimate for it at a further time $Z^{t+k}_{i,j}$. By constructing an observation map $Z^{t}$ which gets updated by replacing observations within the FOV at each time step, the current state of the environment is tracked with a rough estimate for slow progression scenarios or large FOVs.

\textbf{Certainty Factor.} To consider the uncertainty of observation of the past time $t_{obs}$, at the current time $t$, regarding the progress of the fires within the observed area, a certainty value for each cell is defined as follows: 
\begin{equation}
\label{equation: Reliability Factor.}
 c_{i,j}^{t} = 1 - \frac{t - t_{obs}}{t_{max}}
\end{equation}
An element-wise multiplication of the certainty values and the observation matrix obtains an uncertainty-aware observation map. $(\tilde{Z}^t_{i, j} =Z^t_{i, j}c_{i,j}^{t})$. By feeding this compensated map to the value/policy network, the network focuses on the areas with higher certainty and adapts better to highly dynamic cases.

\subsubsection{\textbf{\textit{Belief-Based Representation}}}
Here an alternative to representing the environment state in dynamic scenarios is proposed, in which the input to the value/policy network is the probability of the state being in the ignited states, and is defined as the belief state $(\mathbf{b_{ij}^{t}}  = Pr \{ F_{ij}^{t} = 1  \})$. This probability is assigned by the agent based on a sequence of observations and its initial prior which comes from information about the vegetation type and density.\\

\noindent \textbf{Belief Initialization:}

% It is common to consider conjugate prior distributions for Bayesian belief propagation models. For i.i.d Bernoulli events, a Beta distribution with $\alpha$ and $\beta$ parameters is set based on the initial ignited cell counts in an area representing the spot fire density of the local proximity. However, as the spreading process makes the ignition probabilities of adjacent cells codependent, assuming a Beta distribution for the prior probabilities will no longer hold conjugacy. As a result of both the adjacent co-dependence and the complexity of the underlying phenomena, we can start from any reasonable distribution. Here we simply choose the Beta distribution for which the parameters are initialized before the scanning phase according to the vegetation density and type that the UAV is aware of. While the UAV is in the scanning phase every observation is used to update these priors based on the update rule discussed in (\ref{equation: إBelief Update}).

In Bayesian models for i.i.d Bernoulli events, initial Beta distribution parameters $\alpha$ and $\beta$ are determined by relative success count. However, with fire spread, adjacent cell ignition probabilities become codependent, breaking the Beta distribution's conjugacy. However, we  use a Beta distribution initialized with known vegetation density and type as an improved prior to obtain smoother, faster and more stable convergence (quasi-convergence for highly dynamic cases).

% Moreover, as the fuel consumption rate is a function of the wind and vegetation, each ignition at a time step cannot be fully described by a mono-parametric model such as a Bernoulli distribution and requires joint probability distribution inference approaches such as Bayesian networks to model the co-variation of adjacent cells in a single mode or across modalities. Data-driven spread modeling approaches fall beyond the scope of this. 
\vspace{-0.3cm}
\begin{equation}
\label{equation: إBelief Initialization}
\begin{aligned}
\mathbf{b_{ij}^{0}}  &= \textit{Beta}(\alpha_{ij}, \beta_{ij}), \quad 
\alpha_{ij} = \frac{V_{ij}}{\rho_{ij}}\alpha_{0},\;\beta_{ij} = \frac{\rho_{ij}}{V_{ij}}\beta_{0}
 \end{aligned}
\end{equation}
 
\noindent \textbf{Belief Update:}
The belief update model represents the dynamics of the environment learned by the agent through samples collected at each time step. First, the limitations of Bayesian updates will be discussed and next, a beta-binomial model is proposed to approximate for belief updates.
\vspace{0.2cm} \\
\noindent \textbf{1. Bayesian Update:}
Bayes' rule updates the belief at time $t$, using ignition probabilities and the likelihood of sustained ignition, as shown by the equation for belief updates. The sequential ignition process, allows belief updates based on only ignition and burnout at time $t$.
\begin{equation}
\label{eq: belief update}
\begin{aligned}  
\mathbf{b_{ij}^{t+1}}  &= p(F_{ij}^{t+1} = 1) \\  
\hspace{0.4cm} &=\,  p(F_{ij}^{t+1} = 1 \mid F_{ij}^{t} = 0)(1 - b_{ij}^{t}) \\
&+\, p(F_{ij}^{t+1} = 1 \mid F_{ij}^{t} = 1)b_{ij}^{t} \\ 
\hspace{0.4cm} &=\, \mathbf{(1 - b_{ij}^{t})\,p^{01\,t}_{ij} +\,b_{ij}^{t} (1 - p^{12\,t}_{ij})} \\ 
\end{aligned}
\end{equation} 
% \fa{need to revise the notation}

In Eq. \ref{eq: belief update}, $p_{ij}^{01\,t}$ and $p_{ij}^{12\,t}$  denote the ignition and burnout probability of a cell at time t, respectively. The ignition probability at time $t$ is a weighted sum of the fire spread probability from one cell to another one (Eq. \ref{equation: SpreadingFire}) and the burnout process is a deterministic process that happens as soon as the fuel of a cell finishes.
\begin{align}
\mathbf{p_{ij}^{01\,t}} &= \sum_{i' , j'} p(F_{ij}^{t+1} = 1 \mid  F_{ij}^{t} = 0, F_{i'j'}^{t} = 1){p(f_{i'j'}^{t} = 1)}\\ \nonumber \hspace{0.4cm} &=\, \sum_{i' , j'} b_{i'j'}^{t} \; p(f_{ij}^{t+1} = 1 \mid  f_{ij}^{t} = 0, f_{i'j'}^{t} = 1) \\ \nonumber
\overset{Eq.2}\Rightarrow =\,  &\boxed{\sum_{i' , j'} b_{i'j'}^{t} (\frac{1}{2\sigma^2}e^{-\frac{d^2}{2\sigma^2}}) \left(\frac{W_{||ij}^{t}}{\max(W_{ij}^{t})}\right) \left(1 - \frac{f_{ij}^t}{f_{ij}^0}\right)} 
\end{align}

Considering the perfect wind measurement in ignited cells, the remaining fuel $f_{i, j}^t$, yet remains unknown. Moreover, the burnout likelihood ($p_{ij}^{12\,t}$) of every cell is unknown despite the fuel consumption behavior known by the agent. This is due to the inconsistent monitoring of a cell and the inherent Gaussian noise implemented in the simulated data, accounting for other variables like heat and oxygen flow affecting consumption in reality. Thus, performing the Bayesian update is impossible without approximating wind and fuel measurements at given locations. 
\\
\textbf{2. Heuristic Approach:} An approximation to the Bayesian update for the Beta conjugate prior is to increase alpha and beta by the number of successes and failures of a Bernoulli trial in a group of i.i.d observations respectively. In the equation below, we consider $N$ to be the number of cells observed within a FOV $N = (l_{FoV})^2$. 
\begin{equation}
\label{equation: إBelief Update}
\begin{aligned}
\alpha'  &= \alpha + \sum_{i = 1}^{K} x_i \quad 
\beta' &= \beta + N - \sum_{i = 1}^{K} x_i
 \end{aligned}
\end{equation}

\subsection{\textbf{\textit{Deep Q-Learning}}}

In large state and action spaces, tabular Q-learning is no longer a solution due to large state-action spaces. \cite{mnih2013playing}. A Deep Q-network (DQN) is used in this case to approximate the true Q-values. This DQNs architecture is designed to take the observations/beliefs along with the UAV state parameters in two separate branches and fuse them later on. On one branch, the belief state or the observation  ($z_{t}$ or $ b_{t}$), is fed through a CNN and compressed into an $8 \times 8 \times 256$ feature map after a few layers. Next, it is flattened and passed to a fully connected layer that outputs the latent representation of the environment in a vector of length $16$. On the other branch, the UAV state ($S_{phy}  = (x^{t}, y^{t}, P^{t}, \phi_{U}^{t})$) are fed to three consecutive fully connected layers to reach the same dimension of the latent spatial feature vector.

% \begin{figure}
%     \centering
%     \centerline{\includegraphics[width=0.3\textwidth]{Images/DQN.png}}
%     \caption{DQN Model Architecture}
%     \label{fig:DQN}
% \end{figure}

\vspace{-0.25cm}
\section{Evaluation}

In this section, visual and numerical evaluations of the monitoring goal are presented. For visual evaluation, trajectories of the UAV in static and dynamic environment are shown with trajectories of the observation and belief representation. The experiment parameter values for the discussed results are shown in Table III. 

\subsection{\textbf{Trajectory Analysis.}}

\begin{figure}
    \centering
    \centerline{\includegraphics[width=0.5\textwidth]{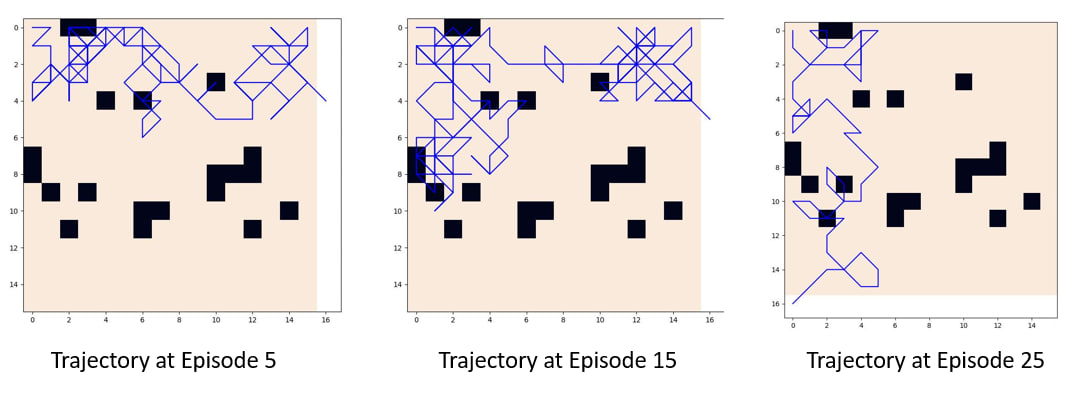}}
    \caption{Trajectories of the UAV in a static environment setting for episodes 5, 15, 25 from left to right. Burnt cells are shown in black. (Trajectory is plotted over the final burnt-out wildfire) }
\label{figure: Trajectory_Static}
\end{figure}

\begin{figure}
    \centering
    \centerline{\includegraphics[width=0.4\textwidth]{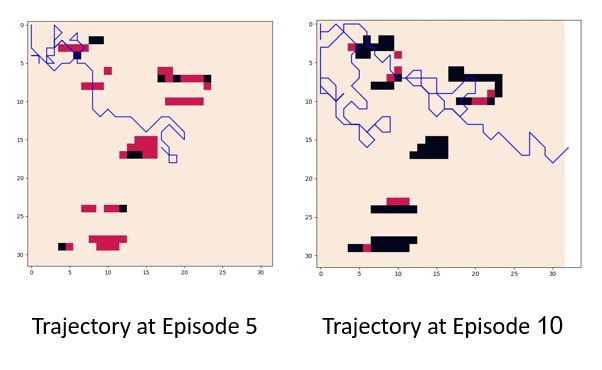}}
    \caption{Bridging over burnt cells - Trajectories of the UAV in a dynamic environment for episodes 5 and 10 from left to right. Burnt cells are shown in black, while ignited cells are shown in red.} 
\label{figure: Trajectory_Dynamic}
\end{figure}

\begin{figure}
    \centering
    \includegraphics[width=0.85\linewidth]{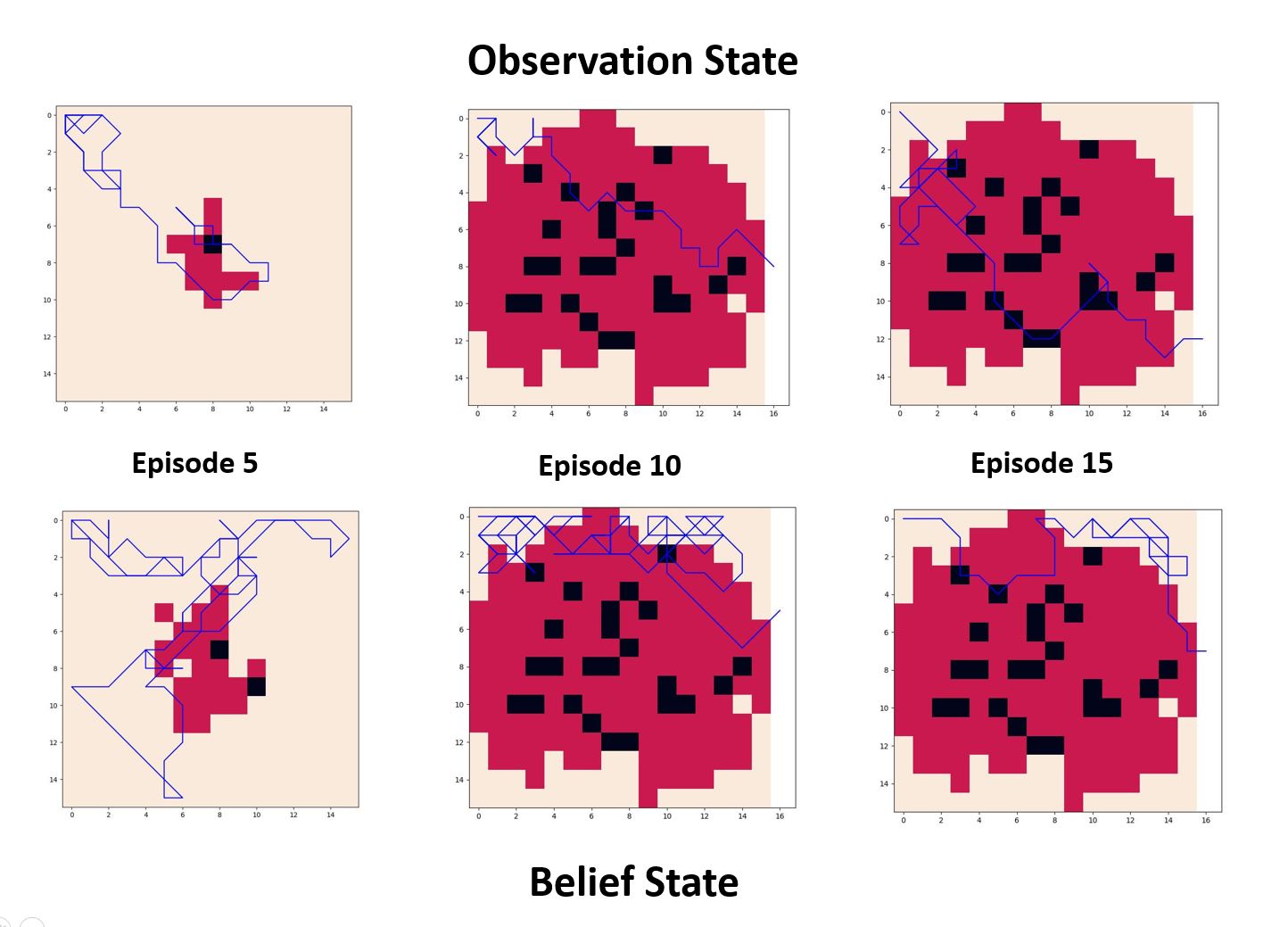}
    \caption{Trajectory of the UAV, which learns in a dynamic radial spread setting with observation and belief maps for a $16 \times 16$ grid. ($3 \times 3$ FOV and constant wind magnitude of $A =0.8 \,A_{max}$)}
  \label{fig:trajectory radial}
\end{figure}

In static fire scenarios (Fig. \ref{figure: Trajectory_Static}), the UAV learns to identify paths with higher q-values, initially exploring more near the take-off area and later expanding its trajectory for broader exploration. In dynamic fire scenarios (Fig. \ref{figure: Trajectory_Dynamic}), within 10 episodes, the UAV adapts to navigate using burnt areas as safe paths in a sparser fire ($32\times32$ grid with $5\times5$ FOV). In the radial fire spread setting (Fig. \ref{fig:trajectory radial}), ($A_{Max} = 100, \sigma_{spr} = 1$), initial strategies using the observation map proved inefficient, but later, using belief states, the UAV learned to identify and use burnt areas as safe passages, optimizing its path in later episodes.

% As shown in Fig.\ref{figure: Trajectory_Dynamic}, even though the agent traces the fire front of two sub-areas in the middle of the map by taking the path in between, many spot fires in the upper right, remain hidden as a result of the burnout penalties. In later episodes, actions that guide flight from frontier cells towards burnt cells are encouraged and acquire a relatively higher value estimate which results in better exploration and high rewards.

\vspace{-0.3cm}

\subsection{\textbf{Coverage and Frontline Tracking Criteria.}}
To compare the UAVs performance in dynamic scenarios for belief v.s. observation state, two criteria are defined. First, the number of detected fire batches across the total grid $(\%Det = \frac{n_{det}}{n_{tot}}$. Second, for monitoring a new criterion called \textit{MIA (Monitored Ignited Area)} is introduced which considers the percentage of ignited area under cover for each batch fire ({$\frac{n_b}{l_{FoV}}$}) and combines it with the normalized minimum distances ($d_b$) to their frontlines. MIA is calculated using Eq. \ref{eq:MIA}.   The results are summarized in Table II. % \ref{table:results_16}
\begin{equation}
\begin{aligned}  
\label{eq:MIA}
\mathbf{MIA} = \underset{{\forall b} \in {B}}{\mathbf{E}}(\frac{n_b}{l_{FoV}^2}\frac{d_b^{max}}{d^{min}_{b}}) &= \underset{{\forall b} \in {B}}{\mathbf{E}}(\frac{n_b}{l_{FoV}^2}\frac{l_{FoV}}{\sqrt{2}d^{min}_{b}}) \\ &= \,\frac{l_{FoV}}{\sqrt{2}}\,\underset{{\forall b} \in {B}}{\mathbf{E}}(\frac{n_b}{d^{min}_{b}}) \leq \frac{l_{FoV}^{3}}{\sqrt{2}}
\end{aligned}
\end{equation} 
 In the equation above, $d_b^{max}$ is half the diagonal of the FOV, which equals the maximum distance of the UAV to an observed ignited cell. $n_{b}$ and $l_{FoV}$ respectively represent the number of covered cells and the size of the field of view

\begin{table}[htbp]
\label{table:results_16}
\caption{Evaluation Results for a $16 \times 16$ grid with a $5\times5$ FOV}
\small % Reduce font size
\centering % Center the table
\setlength{\tabcolsep}{5pt} % Reduce column spacing
\renewcommand{\arraystretch}{1.2} % Increase row height

\begin{tabular}{|c|c|c|c|}
\hline
\textbf{Evaluation} & \textbf{State} & \multicolumn{2}{c|}{\textbf{\textit{Multi-Fire}}} \\
\cline{3-4}
 \textbf{Criteria}&  \textbf{Representation}& {\textit{Static (Batch)}} & {\textit{Dynamic}} \\
\hline

\multirow{2}{*}{Fire Coverage Ratio} & Observation & 86.2\% & 77.3\% \\ \cline{2-4}
 & Belief & 82.5\% & 79.4\% \\
\hline

\multirow{2}{*}{Time-Average MIA} & Observation & {23.25} & {11.91} \\ \cline{2-4}
 & Belief & {18.41} & {14.16} \\
\hline

\end{tabular}
\end{table}

As seen in Table II, in static (slow) scenarios observation states yield a higher coverage ratio and MIA, compared to belief states. This approves expectations as the observations accurately represent the environment after the UAV covers the grid for the first time. In the dynamic case, however, as the observations become outdated, the belief helps the agent assign higher Q-values to actions for tracking the front line, instead of dangerously monitoring it from above.

\begin{table}
\label{tab:experiment}
  \caption{Experimental Settings for Results in Table II} \label{table2}
  \centering 
  \begin{threeparttable}
    \begin{tabular}{cccc}
    %{m{15mm} m{70mm} m{18mm}}
    \textbf{Envir. Settings}  & Value & \textbf{Agent Settings} & Value\\
     \midrule\midrule
    Grid Size  & 16 &  FOV & 5  \\%new row
    \cmidrule(lr){1-2}\cmidrule(lr){3-4} \# Initial Ignitions & 10 &  Steer limit  & $180^{\circ}$  \\ 
    
    \cmidrule(lr){1-2}\cmidrule(lr){3-4} \# Veg. Patches & 5 &  Detection Reward  & 10 \\ 

    \cmidrule(lr){1-2}\cmidrule(lr){3-4} Wind Mag. Var. Period & 20 &  Monitoring Reward  &  10\\ 

    \cmidrule(lr){1-2}\cmidrule(lr){3-4} Wind Phase Var. Period & 80 &  Mvm/Hov Power Ratio  &  2\\ 

    \cmidrule(lr){1-2}\cmidrule(lr){3-4}  Wind Mag. Amp.  & 80 &  Belief Reward  & 40  \\

    \cmidrule(lr){1-2}\cmidrule(lr){3-4} Wind Mag. Var.  Amp. & 20  &  Burnout Penalty  &  -200 \\

    \cmidrule(lr){1-2}\cmidrule(lr){3-4}  Wind Mag. Max & 100 &  Burnout Limit  &  10\\ 
                    
     \midrule\midrule
        Num. Episodes  & 20 & Max. Iterations &  500 \\

    \midrule\midrule
    \end{tabular}
    
\end{threeparttable}
\end{table}

\section{Conclusion}
This paper develops a belief-based DRL solution for UAV path planning in dynamic forest wildfires considering various factors contributing to fire spread including the wind, and vegetation as well as the constraints of low-altitude drones (limited flight time and field of view). The belief-based  state representation in such highly dynamic and partially observable environment where key factors of fire spread are hidden to the UAV with limited sensing and vision capabilities shows promise, by implicitly learning the wildfire spread model through estimating the ignition probability. The belief framework offers a memory-efficient statistic of the POMDP history suitable for low-altitude UAVs.  Moreover, it considers the uncertainty of outdated regions through a certainty factor and offers tunable reward balance between objectives and constraints. Although this approach, may get limited by  multi-modal highly co-variated data which results in complex spatial dependencies, but is capable of adapting to several monitoring tasks and scenarios. 
 It is worth noting that the proposed method focuses on single-agent RL to exhibit state representation potential and future works are encouraged to extend this belief-based model to frameworks with multiple agents such as dec-POMDPs.

% \begin{table}[htbp]
% \label{table:experiment settings}
% \caption{Experiment Settings for Table II }
% \small % Reduce font size
% \centering % Center the table
% \setlength{\tabcolsep}{5pt} % Reduce column spacing
% \renewcommand{\arraystretch}{1.2} % Increase row height

% \begin{tabular}{|c|c|c|c|}
% \hline
% \textbf{} & \textbf{State} & \multicolumn{2}{c|}{\textbf{\textit{Multi-Fire}}} \\
% \cline{3-4}
%  \textbf{Criteria}&  \textbf{Representation}& {\textit{Static (Batch)}} & {\textit{Dynamic}} \\
% \hline

% \multirow{2}{*}{Fire Coverage Ratio} & Observation & 86.2\% & 77.3\% \\
%  & Belief & 82.5\% & 79.4\% \\
% \hline

% \multirow{2}{*}{Time-Average MIA} & Observation & {23.25} & {11.91} \\
%  & Belief & {18.41} & {14.16} \\
% \hline

% \end{tabular}
% \end{table}

\bibliographystyle{ieeetr}
\bibliography{root.bib}
\end{document}